\documentclass[letterpaper, 10 pt, conference]{ieeeconf}  % 
\usepackage{graphicx}
\usepackage{lipsum}
\usepackage{url}
\usepackage{amssymb}
\usepackage{amsmath}
\usepackage{hyperref}
\usepackage{booktabs}

\usepackage[T1]{fontenc} % Use T1 encoding, solve 
\usepackage{dsfont}
\usepackage[table]{xcolor} % in your preamble

\IEEEoverridecommandlockouts                              % 

\title{\LARGE \bf \smash{
StereoAdapter: Adapting Stereo Depth Estimation to Underwater Scenes
}}

\author{
    \textbf{Zhengri Wu}$^{1*}$\quad
    \textbf{Yiran Wang}$^{2*}$\quad 
    \textbf{Yu Wen}$^{1*}$\quad 
    \textbf{Zeyu Zhang}$^{3*\dag}$\quad
    \textbf{Biao Wu}$^{1}$\quad 
    \textbf{Hao Tang}$^{3\ddag}$ \vspace{0.1cm}\\
    $^1$AI Geeks\quad
    $^2$Australian Centre for Robotics\quad
    $^3$Peking University\vspace{0.05cm}\\
    \small $^*$Equal contribution. $^\dag$Project lead.
    $^\ddag$Corresponding author: bjdxtanghao@gmail.com.
}

\begin{document}

\maketitle
\thispagestyle{empty}
\pagestyle{empty}

%%%%%%%%%%%%%%%%%%%%%%%%%%%%%%%%%%%%%%%%%%%%%%%%%%%%%%%%%%%%%%%%%%%%%%%%%%%%%%%%
\begin{abstract}

Underwater stereo depth estimation provides accurate 3D geometry for robotics tasks such as navigation, inspection, and mapping, offering metric depth from low-cost passive cameras while avoiding the scale ambiguity of monocular methods. However, existing approaches face two critical challenges: (i) parameter-efficiently adapting large vision foundation encoders to the underwater domain without extensive labeled data, and (ii) tightly fusing globally coherent but scale-ambiguous monocular priors with locally metric yet photometrically fragile stereo correspondences. To address these challenges, we propose StereoAdapter, a parameter-efficient self-supervised framework that integrates a LoRA-adapted monocular foundation encoder with a recurrent stereo refinement module. We further introduce dynamic LoRA adaptation for efficient rank selection and pre-training on the synthetic UW-StereoDepth-40K dataset to enhance robustness under diverse underwater conditions. Comprehensive evaluations on both simulated and real-world benchmarks show improvements of 6.11\% on TartanAir and 5.12\% on SQUID compared to state-of-the-art methods, while real-world deployment with the BlueROV2 robot further demonstrates the consistent robustness of our approach.
Code: \url{https://github.com/AIGeeksGroup/StereoAdapter}.
Website: \url{https://aigeeksgroup.github.io/StereoAdapter}.
\end{abstract}

\section{Introduction}

Stereo depth estimation is pervasive in robotics for understanding \cite{huang20253dr1,huang2025dc,huang20253d}, navigation \cite{liu2025navr1}, manipulation \cite{song2025maniplvm}, and inspection \cite{song2025hazards}, offering \emph{metric} 3D from low-cost passive binocular cameras and avoiding the scale ambiguity that plagues monocular depth \cite{HartleyZisserman2004,Geiger2012KITTI}. Underwater depth estimation is equally critical for AUV/ROV mapping, infrastructure inspection (e.g., pipelines and hulls), ecology monitoring, and archaeology, where reliable geometry directly impacts autonomy and safety \cite{Zhang2022UnderwaterSLAM,Nauert2023InterventionAUVs}.
Recent efforts connect monocular priors with stereo geometry from different angles: TiO-Depth unifies \emph{monocular} and \emph{binocular} self-supervised depth estimation within a single two-in-one framework \cite{zhou2023two}, whereas Stereo Anywhere couples stereo geometry with robust priors from monocular vision foundation models (VFMs) to attain strong zero-shot generalization even in textureless, specular, or transparent scenes \cite{bartolomei2025stereo}. However, underwater imaging violates photometric assumptions commonly exploited by terrestrial stereo due to wavelength-dependent attenuation, forward and backscattering, and refraction at water–glass interfaces, which induces a severe domain shift \cite{akkaynak2018revised} and leaves two key challenges: (i) \emph{parameter-efficiently} adapting large foundation encoders to the underwater domain without extensive labels, and (ii) \emph{tightly fusing} globally coherent yet scale-ambiguous monocular priors with locally metric but photometrically fragile stereo correspondences within a self-supervised pipeline.

Our motivation is to reconcile the complementary strengths of monocular vision foundation models and stereo geometry while keeping adaptation economical in both labels and parameters. Concretely, we (i) \emph{adapt rather than replace} a strong foundation encoder to underwater appearance using lightweight low-rank modules instead of full fine-tuning, and (ii) treat the resulting monocular prior as \emph{initialization and guidance} for an iterative stereo matcher so that local metric constraints can correct global-scale or texture-induced errors under a purely self-supervised objective. This perspective yields a pipeline that is robust to severe underwater domain shift and practical for deployment.

\begin{figure}[t]
    \centering
    \includegraphics[width=\columnwidth]{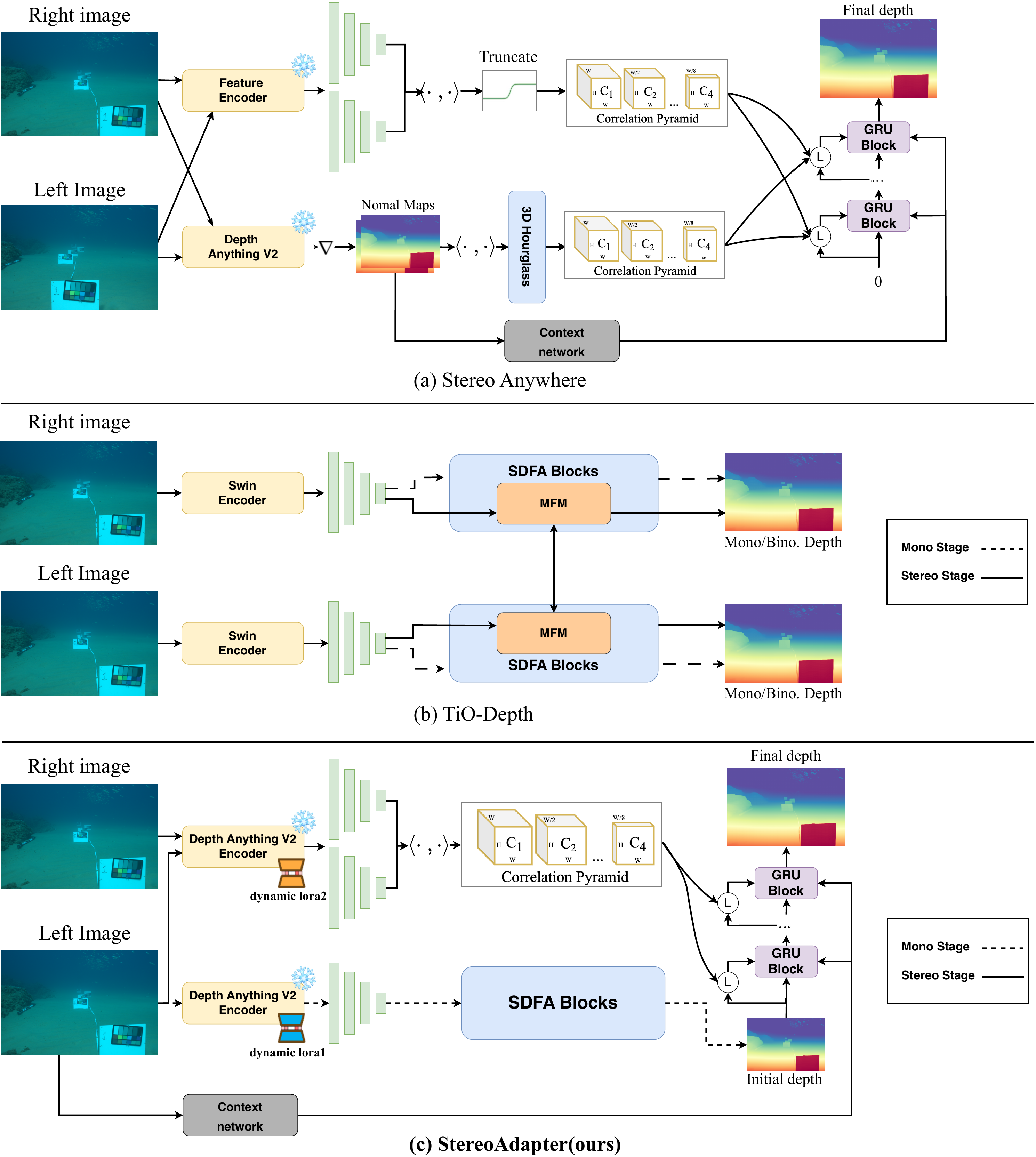}
    \caption{\textbf{Comparison between our methods and other baselines:} (a) TiO-Depth (b) Stereo Anywhere (c) StereoAdapter (Ours).}
    \label{fig:main-arch}
\end{figure}

To address the above challenges, we propose \textbf{StereoAdapter}, a parameter-efficient self-supervised framework that merges a LoRA-adapted monocular foundation encoder with a recurrent stereo refinement module. \emph{Architecturally}, StereoAdapter exposes multi-scale features from a Depth-Anything-style encoder (adapted with LoRA) to produce a coarse, globally consistent disparity prior, and fuses it with a multi-scale correlation pyramid inside a GRU-style updater to recover locally metric depth at full resolution, as shown in Figure \ref{fig:main-arch} (c). \emph{Training-wise}, we employ dynamic LoRA to automatically select effective ranks per layer and consolidate the surviving components into the base weights, and we optimize with monocular-prior guidance, photometric reconstruction, occlusion-aware masking, and edge-aware smoothness enabling continual adaptation without dense underwater ground truth. \emph{On the data side}, we introduce \textbf{UW-StereoDepth-40K} for pre-training, synthesized with high-fidelity underwater stereo in UE5, ensuring robust cross-view consistency while varying attenuation, scattering, particles, and baselines to emulate diverse ROV setups.
Furthermore, we conducted comprehensive experiments on both simulated and real-world underwater stereo depth benchmarks, including TartanAir \cite{wang2020tartanair} and SQUID \cite{berman2020underwater}, achieving improvements of 6.11\% and 5.12\% in RMSE compared to the state-of-the-art methods. We also performed real-world evaluations with the BlueROV2 robot, which demonstrated consistently robust performance.

The main contribution of our work can be summarized as follows:

\begin{itemize}
    \item We propose \textbf{StereoAdapter}, a parameter-efficient self-supervised framework that integrates a LoRA-adapted monocular foundation encoder with a recurrent stereo refinement module.
    \item We design a training strategy that leverages \textit{dynamic LoRA adaptation} for efficient rank selection and pre-training on the synthetic \textbf{UW-StereoDepth-40K} dataset to enhance underwater robustness.
    \item We conduct comprehensive experiments on both simulated and real-world benchmarks including TartanAir~\cite{wang2020tartanair} and SQUID~\cite{berman2020underwater}, as well as real-world evaluations with the BlueROV2 robot, demonstrating consistent improvements over state-of-the-art methods.
\end{itemize}

\section{Related Work}

\paragraph{Deep Stereo Matching and Self-supervised Learning}

Deep learning has revolutionized stereo matching through end-to-end architectures. Early methods like GC-Net \cite{kendall2017end}, PSMNet \cite{chang2018pyramid}, and GA-Net \cite{zhang2019ga} pioneered cost volume regularization with 3D convolutions. The field shifted toward iterative optimization with RAFT-Stereo \cite{lipson2021raft}, which inspired subsequent works including IGEV-Stereo \cite{xu2023iterative} and Selective-IGEV \cite{wang2024selective}. Recent approaches explored Transformer architectures \cite{li2021revisiting} and self-supervised pretraining \cite{weinzaepfel2023croco}.

The pursuit of robust generalization across diverse domains has led to the emergence of foundation models for depth estimation. MiDaS \cite{ranftl2020towards} and DPT \cite{ranftl2021vision} demonstrated that training on mixed datasets enables remarkable zero-shot transfer capabilities. Depth Anything \cite{yang2024depthanything} unleashed the potential of large-scale unlabeled data, while Depth Anything V2 \cite{yang2024depthanythingv2} further enhanced fine-grained details and robustness through synthetic-to-real knowledge transfer. These monocular foundation models have recently been integrated into stereo matching frameworks: Stereo Anywhere \cite{bartolomei2025stereo} achieves robust zero-shot stereo matching by combining monocular depth priors with stereo geometry, particularly excelling in scenarios where traditional stereo or monocular methods fail independently. FoundationStereo \cite{wen2025foundation} adapts foundation model architectures specifically for stereo matching tasks, while DEFOM-Stereo \cite{jiang2025defom} introduces dynamic fusion mechanisms to optimally combine monocular and stereo cues based on scene characteristics.

The scarcity of ground-truth depth annotations has also motivated self-supervised learning approaches. For monocular estimation, Zhou et al. \cite{zhou2017unsupervised} introduced learning from video sequences through ego-motion estimation, while Monodepth \cite{godard2017unsupervised} and Monodepth2 \cite{godard2019digging} leveraged left-right consistency in stereo pairs. Advanced techniques include depth hints \cite{watson2019self}, reversing distillation cycles between monocular and stereo networks \cite{aleotti2020reversing}, and revealing reciprocal relations between self-supervised stereo and monocular learning \cite{chen2021revealing}. TiO-Depth \cite{zhou2023two} unified monocular and binocular tasks in a single framework, demonstrating their complementary nature. These self-supervised methods provide valuable alternatives when labeled data is scarce, particularly important for challenging domains like underwater environments.

\paragraph{Domain Adaptation for Stereo Matching}

Despite advances in stereo matching, cross-domain generalization remains challenging \cite{zhang2020domain,cai2020matching}. While methods like masked representation learning \cite{rao2023masked} show promise for terrestrial scenes, they struggle with extreme domain shifts. Existing datasets primarily focus on terrestrial environments: synthetic datasets (Scene Flow \cite{mayer2016large}, HyperSim \cite{roberts2021hypersim}) and real benchmarks (KITTI \cite{geiger2012we}, Middlebury \cite{scharstein2014high}, Booster \cite{zamaramirez2022booster}) provide comprehensive coverage but cannot capture underwater challenges.

Parameter-efficient fine-tuning offers a promising solution for domain adaptation with limited data. LoRA \cite{hu2021lora} and its variants \cite{dettmers2023qlora,liu2024dora,zhang2023adalora,he2022sparseadapter} demonstrated that large models can be adapted with minimal trainable parameters, particularly effective when target domain data is scarce. Recent vision applications \cite{zhang2024dares,chen2022adaptformer,chang2023revisiting} show that adaptive parameter distribution improves adaptation efficiency in challenging domains.

\paragraph{Underwater Depth Estimation}

Underwater imaging violates fundamental assumptions of standard vision algorithms due to wavelength-dependent attenuation, backscattering, and refraction \cite{akkaynak2018revised}. Early underwater datasets \cite{berman2020underwater,islam2020semantic} lacked stereo depth annotations. UW-Stereo \cite{lv2025uwstereo} provided 29,568 synthetic stereo pairs, but the sim-to-real gap persists.

Recent underwater depth estimation methods address these challenges through various strategies. UWStereo \cite{ye2023underwater} proposed comprehensive adaptation with style, semantic, and disparity modules. UWNet and Fast-UWNet \cite{uwnet2024} developed attention mechanisms for real-time underwater processing. However, these methods require extensive underwater data or complex adaptation pipelines.

The combination of severe domain shift and data scarcity in underwater scenarios motivates our approach: leveraging foundation models' robustness, self-supervised learning's adaptability, and parameter-efficient fine-tuning through LoRA \cite{dynamiclora2024}. This enables effective underwater depth estimation by adapting pre-trained models with minimal underwater-specific data, bridging the gap between terrestrial knowledge and underwater applications.

\begin{figure*}[t]
    \centering
    \includegraphics[width=\textwidth]{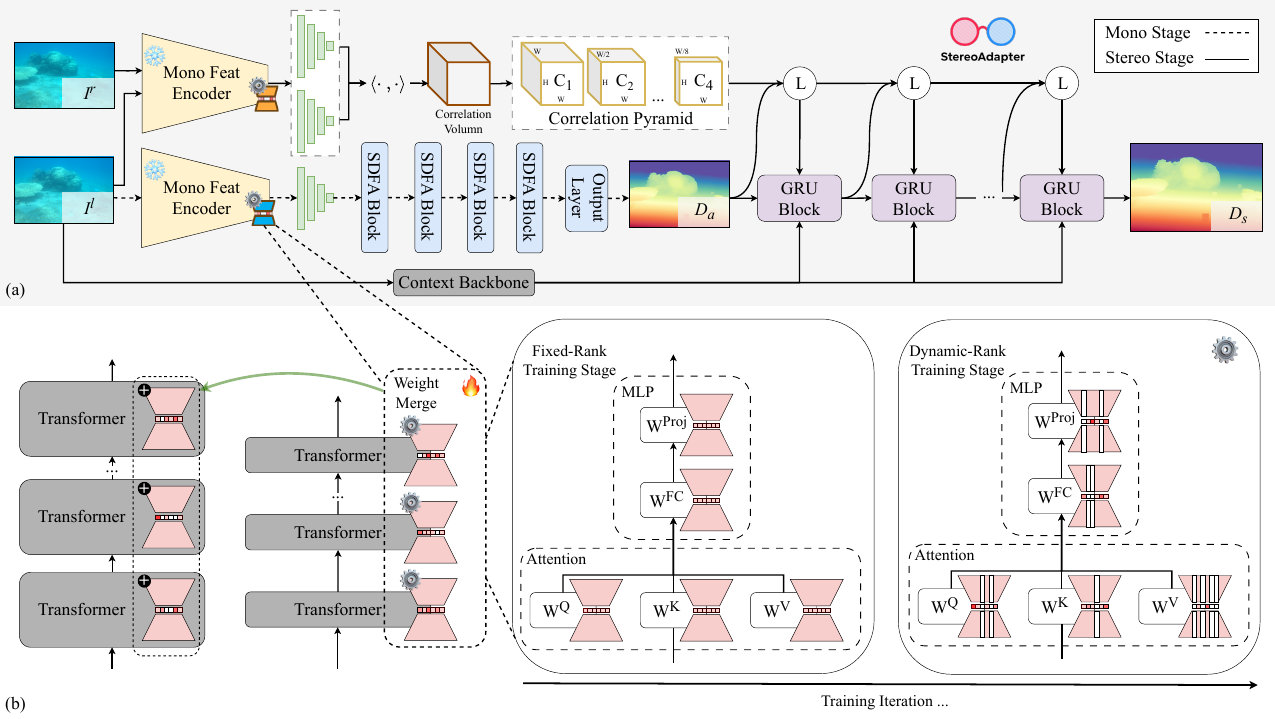}
    \caption{\textbf{Detailed architecture of the StereoAdapter:} (a) two-stage self-supervised training pipeline; (b) update mechanism with LoRA.}
    \label{fig:arch}
\end{figure*}

\section{The Proposed Method}

\subsection{Overview}
We proposed StereoAdapter, which adopts a self-supervised learning pipeline for underwater stereo depth prediction which leverages monocular depth estimation to guide stereo depth prediction as shown in Fig \ref{fig:arch} (a). The framework consists of two main training stages. In the first stage, we employ a depth foundation model, Depth Anything V2 \cite{yang2024depth}, for monocular underwater depth prediction. We utilize LoRA \cite{hu2021lora} to adapt the encoder to underwater scenes, and then decode the feature pyramid output from the encoder to obtain scene disparity. In the second stage, the coarse-grained disparity prediction from the first stage is used for disparity initialization and fused with the cost volume pyramid in the recurrent module. Through progressive iterations, we generate accurate fine-grained disparities and convert them to depth.
Furthermore, inspired by \cite{dynamiclora2024}, we design the dynamic LoRA for continuous learning among different domains, enabling the encoder from VFM to better collaborate with stereo disparity prediction, as shown in Fig \ref{fig:arch} (b).

\subsection{Architecture}

\paragraph{Monocular Depth Estimation}
In the monocular depth estimation pathway, we leverage a pretrained Depth Anything V2 \cite{yang2024depth} model. We directly utilize the multi-scale feature pyramids $\{F_i\}_{i=1}^4$ at resolutions $\{H/4, H/8, H/16, H/32\}$ from the DPT's reassemble modules. To adapt these pretrained representations to underwater scenes, we incorporate LoRA \cite{dynamiclora2024} modules in the transformer encoder, enabling efficient domain adaptation while preserving the learned geometric priors. This approach allows us to benefit from Depth Anything V2's strong depth representations while adapting to underwater-specific characteristics with minimal additional parameters.

The adapted features are then processed through SDFA blocks \cite{yang2022monovit} for progressive aggregation. Each SDFA block combines features from adjacent scales, maintaining spatial coherence while incorporating multi-scale context. The decoder produces a discrete disparity volume $V \in \mathbb{R}^{N \times H \times W}$, where $N$ represents the number of predefined disparity levels. 

During training, we generate the primary volume $V_m$ through our decoder branches. This volume is directly used for photometric reconstruction following \cite{zhou2023two}. Specifically, we employ the discrete depth constraint to generate a reconstructed image $\hat{I}_l$ from the primary volume $V_m^r$ of the right view and the original right image $I_r$. The monocular training objective consists of three components: a photometric reconstruction loss $\mathcal{L}_{rec}^{mono}$ measuring the difference between $\hat{I}_l$ and $I_l$, a multi-scale edge-aware smoothness loss $\mathcal{L}_{smooth}^{mono}$:
\begin{equation}
\mathcal{L}_{mono} = \mathcal{L}_{rec}^{mono} + \lambda_1 \mathcal{L}_{smooth}^{mono},
\end{equation}
where $\lambda_1$ is predefined weighting parameters. During this stage, only the LoRA parameters and decoder components are optimized while keeping the pretrained encoder weights frozen.

\paragraph{Correlation Pyramids Building}

We construct multi-scale correlation pyramids from stereo features while leveraging our mono stage's scaled predictions and sparse stereo matching for robust metric depth estimation. Our mono stage directly produces metric-scaled depth, which we further refine using sparse correspondences for improved local accuracy.

\textbf{Stereo Correlation Pyramid.}
Given feature maps $\mathbf{f}_L, \mathbf{f}_R \in \mathbb{R}^{C \times H/4 \times W/4}$ extracted from the stereo pair, we compute the 4D correlation volume through the inner product:
\begin{equation}
    \mathbf{C}(i,j,d) = \langle \mathbf{f}_L(i,j), \mathbf{f}_R(i,j-d) \rangle, \quad d \in [0, D_{\max}].
\end{equation}
We then construct a multi-scale pyramid $\{\mathbf{C}^{(l)}\}_{l=0}^{3}$ via average pooling, where each level provides correlation features at different granularities for coarse-to-fine refinement.

\textbf{Hybrid Scale Alignment and Refinement.}
Our mono stage produces metric-scaled depths $\mathbf{M}_L^{\text{mono}}, \mathbf{M}_R^{\text{mono}}$ that already possess global scale consistency. To ensure metric accuracy and improve local precision, we employ sparse stereo correspondences as verification and refinement anchors. Through feature matching with bidirectional consistency check, we obtain sparse metric depth measurements:
\begin{equation}
    \mathbf{D}_{\text{sparse}}(p) = \frac{f \cdot b}{d_p}, \quad p \in \mathcal{P}_{\text{matched}}.
\end{equation}

We first verify the mono stage's scale by comparing with sparse depths:
\begin{equation}
    \alpha = \frac{1}{|\mathcal{P}_{\text{matched}}|} \sum_{p \in \mathcal{P}_{\text{matched}}} \frac{\mathbf{D}_{\text{sparse}}(p)}{\mathbf{M}_L^{\text{mono}}(p)}.
\end{equation}
If $|\alpha - 1| < \tau$ (typically $\tau = 0.1$), the mono stage scale is considered reliable. Otherwise, we compute the corrective scale and shift:
\begin{equation}
    \min_{\hat{s}, \hat{t}} \sum_{p \in \mathcal{P}_{\text{matched}}} w_p \left\| (\hat{s} \cdot \mathbf{M}_L^{\text{mono}}(p) + \hat{t}) - \mathbf{D}_{\text{sparse}}(p) \right\|^2.
\end{equation}

The initially aligned depths are:
\begin{equation}
    \hat{\mathbf{M}}_L^{(0)} = \begin{cases}
        \mathbf{M}_L^{\text{mono}}, & \text{if } |\alpha - 1| < \tau \\
        \hat{s} \cdot \mathbf{M}_L^{\text{mono}} + \hat{t}, & \text{otherwise}
    \end{cases}
\end{equation}

Subsequently, we refine these depths through confidence-weighted propagation of local corrections:
\begin{equation}
    \hat{\mathbf{M}}_L(p) = \hat{\mathbf{M}}_L^{(0)}(p) + \sum_{q \in \mathcal{P}_{\text{matched}}} w_{pq} \cdot (\mathbf{D}_{\text{sparse}}(q) - \hat{\mathbf{M}}_L^{(0)}(q)),
\end{equation}
where bilateral weights $w_{pq}$ preserve edges while propagating corrections:
\begin{equation}
    w_{pq} = \exp\left(-\frac{\|p-q\|^2}{2\sigma_d^2}\right) \cdot \exp\left(-\frac{\|\mathbf{I}(p)-\mathbf{I}(q)\|^2}{2\sigma_c^2}\right).
\end{equation}
This hybrid approach leverages the global consistency of our mono stage while ensuring local metric accuracy through sparse anchors.

\paragraph{Stereo Depth Estimation}

Our stereo depth estimation module refines the initial metric-scaled predictions from the mono stage through iterative optimization using stereo correspondences. We adopt a recurrent refinement framework that leverages both enhanced context features and stereo matching cues.

\textbf{Combined Context Encoder.}
Inspired by DEFOM-Stereo \cite{jiang2025defom}, we directly leverage the VFM encoder from our monocular depth estimation stage.It has already been fine-tuned with fused weights for underwater scene understanding. This pre-adapted encoder provides robust feature representations that capture both general visual semantics and domain-specific characteristics. For each resolution level $l \in \{4, 8, 16\}$, we extract hierarchical features from the DPT's Reassemble modules:
\begin{equation}
    \mathbf{h}_{\text{ViT}}^{(l)} = \text{Reassemble}_l^{\text{LoRA}}(\mathbf{I}_L),
\end{equation}
where $\text{Reassemble}_l^{\text{LoRA}}$ denotes the Reassemble module with fused LoRA weights from the mono stage.

In parallel, we employ a lightweight CNN-based context encoder to extract complementary local features following :
\begin{equation}
    \mathbf{h}_{\text{CNN}}^{(l)} = \text{CNN}_l(\mathbf{I}_L).
\end{equation}

To effectively combine these features, we apply learnable convolutional blocks for channel alignment and perform element-wise addition:
\begin{equation}
    \mathbf{h}^{(l)} = \text{Conv}_{\text{align}}(\mathbf{h}_{\text{ViT}}^{(l)}) + \mathbf{h}_{\text{CNN}}^{(l)}.
\end{equation}

This design efficiently reuses the domain-adapted representations from our mono stage while augmenting them with CNN features for enhanced spatial precision. The resulting multi-scale context features $\{\mathbf{h}^{(l)}\}_{l \in \{4,8,16\}}$ initialize the GRU's hidden state and are integrated at each iteration to provide consistent geometric and semantic guidance throughout the refinement process.

\textbf{Iterative Disparity Refinement.}
We aim to estimate a series of refined disparity maps $\{\mathbf{d}^{(1)} = f \cdot b/\hat{\mathbf{M}}_L, \mathbf{d}^{(2)}, \ldots, \mathbf{d}^{(l)}, \ldots\}$ exploiting the guidance from our combined context encoder and stereo correlation pyramid. Starting from the GRU update operator, we employ a lookup operator that extracts correlation features $\mathbf{c}^{(l)}$ from the multi-scale correlation pyramid $\{\mathbf{C}^{(l)}\}_{l=0}^{3}$. 

The correlation features $\mathbf{c}^{(l)}$ are processed by a two-layer encoder and concatenated with features derived from the current disparity estimation $\mathbf{d}^{(l)}$ and the multi-scale context features $\mathbf{g}^{(l)}$ from our combined encoder. This concatenation is further processed by a 2D convolutional layer, and then fed to the ConvGRU operator:
\begin{equation}
    \mathbf{h}^{(l+1)}, \Delta\mathbf{d}^{(l)} = \text{ConvGRU}(\mathbf{h}^{(l)}, [\mathbf{c}^{(l)}, \mathbf{g}^{(l)}, \phi(\mathbf{d}^{(l)})]),
\end{equation}
where $\phi(\cdot)$ denotes feature extraction from the current disparity map.

We adopt the same upsampling module from \cite{lipson2021raft} to upsample the final disparity to full resolution and then project it back to depth, producing metric-accurate depth estimates suitable for underwater robotics applications.

\textbf{Joint training strategy.}
We employ a comprehensive loss function that leverages monocular predictions as strong priors while enforcing stereo consistency. Let $\mathbf{d}^{(L)}$ denote the final disparity after all iterations.

The stereo reconstruction loss enforces photometric consistency between the left image and the warped right image:
\begin{equation}
    \mathcal{L}_{\text{rec}} = \alpha \|\tilde{\mathbf{I}}_L - \mathbf{I}'_L\|_1 + (1-\alpha)\text{SSIM}(\tilde{\mathbf{I}}_L, \mathbf{I}'_L),
\end{equation}
where $\tilde{\mathbf{I}}_L$ is reconstructed by warping $\mathbf{I}_R$ using the predicted disparity $\mathbf{d}^{(L)}$.

To handle occlusions, we leverage the monocular predictions:
\begin{equation}
    \mathbf{I}'_L = \mathbf{M}_{\text{occ}} \odot \mathbf{I}_L + (1 - \mathbf{M}_{\text{occ}}) \odot \tilde{\mathbf{I}}_L^{\text{mono}},
\end{equation}
where $\mathbf{M}_{\text{occ}}$ is the occlusion mask computed from the initial monocular disparity $\mathbf{d}^{(1)} = f \cdot b/\hat{\mathbf{M}}_L$, and $\tilde{\mathbf{I}}_L^{\text{mono}}$ is reconstructed using monocular depth.

The disparity guidance loss leverages monocular predictions to regularize stereo refinement:
\begin{equation}
\begin{aligned}
    \mathcal{L}_{\text{guide}} = &\|\nabla_x \mathbf{d}^{(1)} - \nabla_x \mathbf{d}^{(L)}\|_1 + \|\nabla_y \mathbf{d}^{(1)} - \nabla_y \mathbf{d}^{(L)}\|_1 \\
    &+ \mathbf{M}_{\text{out}} \odot \|\mathbf{d}^{(1)} - \mathbf{d}^{(L)}\|_1,
\end{aligned}
\end{equation}
where $\mathbf{d}^{(1)}$ and $\mathbf{d}^{(L)}$ represent the initial monocular and final stereo disparity maps respectively, $\nabla_x$ and $\nabla_y$ denote the horizontal and vertical gradient operators, and $\mathbf{M}_{\text{out}}$ masks pixels with invalid reprojections.

Additionally, we apply an edge-aware smoothness loss $\mathcal{L}_{\text{smooth}}^{\text{stereo}}$ for final prediction.
The complete stereo training objective is:
\begin{equation}
    \mathcal{L}_{\text{stereo}} = \mathcal{L}_{\text{rec}}^{\text{stereo}} + \lambda_3 \mathcal{L}_{\text{smooth}}^{\text{stereo}} + \lambda_4 \mathcal{L}_{\text{guide}},
\end{equation}
where $\{\lambda_3, \lambda_4\}$ are weighting parameters. During this stage, a new set of LoRA weights will be learned for adapting the VFM encoder from a monocular to a stereo setup.

\subsection{Dynamic LoRA}
\textbf{Adaptive Rank Selection Mechanism.} To address the substantial domain gap between terrestrial pre-training environments and underwater deployment scenarios, we incorporate Low-Rank Adaptation (LoRA) \cite{hu2021lora} for efficient encoder fine-tuning. For a weight matrix $\textbf{W}_0 \in \mathds{R}^{d \times k}$, Traditional LoRA introduces low-rank decomposition matrices $\textbf{B} \in \mathds{R}^{d \times r}$ and $\textbf{A} \in \mathds{R}^{r \times k}$ to approximate weight updates:
\begin{equation} % LoRA
    h = \textbf{W}_0 \textbf{x} + \Delta{\textbf{W} \textbf{x}} = \textbf{W}_0 \textbf{x} + \textbf{B} \textbf{A} \textbf{x},
\end{equation}
where $r \ll (\text{min}(d, k))$ represents the adaptation rank.

However, fixed-rank approaches fail to account for the varying adaptation requirements across different network components and layers. This limitation often leads to either insufficient adaptation capacity with overly conservative ranks or excessive parameter overhead with unnecessarily high ranks. Inspired by \cite{dynamiclora2024}, we design a dynamic LoRA that enables adaptive rank optimization based on the importance of individual singular components to address this limitation.

The dynamic LoRA reformulates the low-rank update through a learnable importance weighting mechanism. Instead of fixed-rank decomposition, it introduces learnable importance weights $\textbf{w} \in \mathds{R}^r$ that modulate the contribution of each rank component:
\begin{equation} % CoDyRA weight
    \Delta{\textbf{W}^{t, m}} = \sum_{i=1}^{r} \textbf{w}_i^{t, m} \textbf{B}_i^{t, m} \textbf{A}_i^{t, m},
\end{equation}
where $\textbf{w}_i^{t, m}$ represents the learned importance of the $i$-th rank component, and $\textbf{B}_i^{t, m}$, $\textbf{A}_i^{t, m}$ correspond to the $i$-th singular vectors of the low-rank decomposition. This formulation draws inspiration from Singular Value Decomposition (SVD) \cite{zhang2023adalora, dynamiclora2024, ding2023sparse, liu2024alora, meng2024pissa, zhang2024milora}, where importance weights correspond to learnable approximations of the diagonal elements in the singular value matrix, enabling the model to dynamically emphasize the most relevant subspace directions for the current adaptation task.

\textbf{Sparsity-Regularized Optimization.} The importance weights $\mathbf{w}$ are initialized randomly and optimized jointly with the low-rank matrices through gradient descent. To ensure only the most impactful rank components are retained while eliminating redundant parameters, our approach incorporates $\ell1$ sparsity regularization:
\begin{equation} % CoDyRA loss
    \mathcal{L}_{train}^t = \mathcal{L}_{sup}^t + \lambda \sum_{m = 1}^M \|\textbf{w}^{t, m}\|_1,
\end{equation}
where $\mathcal{L}_{sup}$ denotes the supervised learning objective, and $\lambda$ controls the regularization strength. This sparsity constraint promotes automatic rank selection by driving less important weights toward zero, performing continuous rank pruning during optimization.

The non-differentiable nature of $\ell 1$ regularization necessitates specialized optimization techniques. Rather than directly applying this $\ell 1$ regularization during optimization, the dynamic LoRA employs the proximal gradient method with soft-thresholding operations to handle the sparsity constraint. The weight update rule for importance parameters follows:
\begin{equation} % Soft threshold
    \textbf{w}_i^{t, m} := \mathds{1} (\left| \hat{\textbf{w}}_i^{t, m} \right| > \kappa) \cdot (\hat{\textbf{w}}_i^{t, m} + \text{sign}(\hat{\textbf{w}}_i^{t, m}) \cdot \kappa),
\end{equation}
where $\hat{\textbf{w}}_i^{t, m}$ denotes the importance weight after gradient update from the supervised loss, $\kappa$ is the threshold parameter, and $\mathds{1}(\cdot)$ is the indicator function. The threshold parameter $\kappa$ is gradually increased during training from zero to a maximum value $\kappa_{max}$, allowing initial exploration of all rank components before applying progressively stronger sparsity constraints.

Additionally, in accordance with the X-TAIL experimental protocol, we employ a two-stage training procedure. The dense training stage constitutes 50\% of the total iterations during which the soft-thresholding operation is not applied, followed by the sparse training stage where the soft-thresholding mechanism is activated to perform rank pruning. The dense training stage is essential as it allows all rank components to capture task-relevant information before sparsity constraints are imposed, preventing premature elimination of potentially important adaptation directions.

\textbf{Continual Weight Integration.} Upon completion of each adaptation phase, ranked components with non-zero importance weights are consolidated into the base network parameters. This integration process eliminates the computational overhead associated with auxiliary adaptation modules during inference while preserving the acquired domain-specific knowledge. The weight merging operation follows:
\begin{equation}
    \hat{\textbf{W}} = \textbf{W}_0 + \Delta{\textbf{W}} = \textbf{W}_0 + \sum_{i=1}^{r} \textbf{w}_i \textbf{B}_i \textbf{A}_i,
\end{equation}
where only rank components with significant importance weights contribute to the final parameter update.

\subsection{Data Synthesis}
\begin{figure}[t]
    \centering
    \includegraphics[width=1.0\columnwidth]{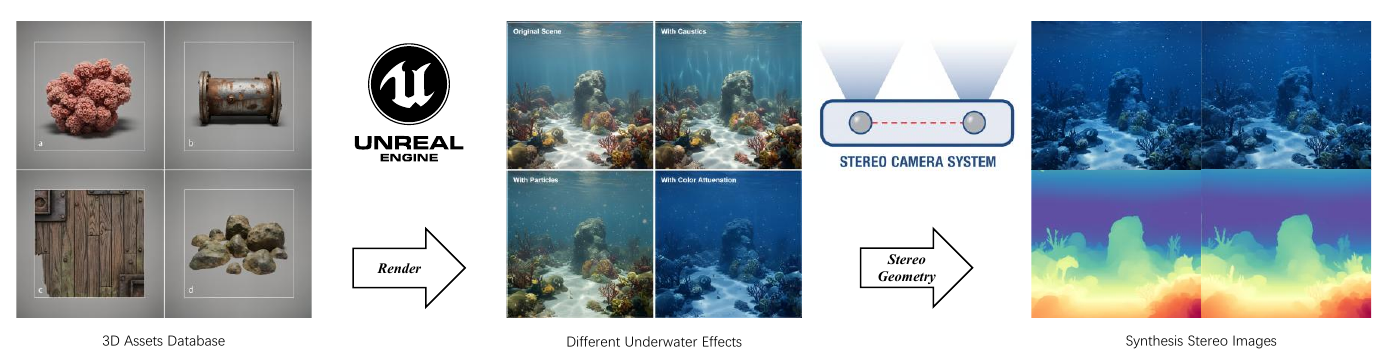}
    \caption{\textbf{Data synthesis pipeline.} Unreal Engine 5 rendering pipeline for UW-StereoDepth-40K dataset.}
    \label{fig:data}
\end{figure}

Inspired by UWstereo~\cite{lv2025uwstereo}, we construct a large-scale synthetic underwater stereo dataset, termed UW-StereoDepth-40K, using Unreal Engine 5 (UE5). UE5's advanced rendering capabilities enable physically accurate simulation of underwater visual effects while maintaining precise geometric consistency required for stereo matching applications. The overall data generation pipeline is illustrated in Figure \ref{fig:data}.

\paragraph{Rendering Pipeline}
We leverage UE5's powerful ray tracing and global illumination systems to create photorealistic underwater environments. Our pipeline encompasses four distinct underwater scenes—coral reefs, industrial structures, shipwrecks, and natural seabed environments—populated with diverse 3D assets. These virtual resources include high-resolution scanned corals, detailed marine flora, underwater robots, submerged vehicles, and various marine structures collected from professional 3D asset libraries and photogrammetry scans.

The key advantage of using UE5 over generative approaches lies in its ability to maintain robust stereo consistency. Unlike diffusion-based methods that introduce stochastic variations between views, UE5 renders both left and right images from precisely calibrated virtual stereo cameras, ensuring pixel-aligned geometric relationships and consistent appearance across the stereo pair. This eliminates the correspondence ambiguity issues inherent in generative approaches while preserving the photorealism necessary for effective domain transfer.

\paragraph{Environmental Variation}
To enhance dataset diversity and improve model generalization, we systematically vary multiple environmental parameters. Camera baselines are sampled from \{4cm, 10cm, 20cm, 40cm\} to accommodate the diverse baseline configurations found across different underwater ROV platforms. This range covers typical stereo configurations from compact inspection ROVs with narrow baselines to larger survey vehicles with wider stereo separation, ensuring our trained models can generalize across various underwater robotic systems.

Additionally, we introduce realistic underwater effects including caustic patterns generated through water surface simulation, floating particles with physics-based motion, and depth-dependent color attenuation. These environmental variations ensure our synthetic data captures the full spectrum of challenges present in real underwater scenarios.

\paragraph{Dataset Generation and Quality Control}
The UW-StereoDepth-40K dataset is generated by processing continuous camera trajectories through the virtual environments, extracting stereo pairs at regular intervals. We implement automatic filtering to remove frames with insufficient texture information and frames with extreme depth distributions that exceed typical underwater robotic operation ranges (>50m).

Each stereo pair is rendered at $1280 \times 960$ resolution with accompanying dense depth ground truth and semantic segmentation masks. Quality assurance involves both automated metrics and manual inspection. We compute structural similarity indices between stereo pairs to ensure consistency, and employ domain experts to evaluate photorealism and identify potential artifacts. The final dataset comprises 40,000 high-quality stereo image pairs covering diverse underwater scenarios, providing comprehensive training data for robust underwater stereo matching models.

\section{Experiments}

\subsection{Datasets and Evaluation Metrics}
\paragraph{Datasets}
For training, we utilize our synthetic dataset UW-StereoDepth-40K, generated using Unreal Engine 5, which consists of 40,000 stereo image pairs from various underwater scenes. For evaluation, we conduct experiments on two real-world underwater stereo datasets. The first is a subset of TartanAir \cite{wang2020tartanair} containing 13,583 underwater stereo image pairs from 22 different sequences. The second is the SQUID dataset \cite{berman2020underwater}, which comprises 57 stereo pairs captured from four distinct scenes.

\paragraph{Evaluation metrics}
We adopt standard depth estimation metrics to comprehensively evaluate our method. Following established protocols, we report Relative Error (REL), Squared Relative Error (SQ REL), Root Mean Square Error (RMSE), and Log Root Mean Square Error (LOG RMSE) to assess accuracy. Additionally, we compute threshold accuracy metrics $\delta < 1.25^i$ (denoted as A1, A2, A3 for $i = 1, 2, 3$ respectively) to measure the percentage of pixels with depth predictions within specified error thresholds.

\begin{table*}[t]
\centering

\caption{Evaluation on the TartanAir Underwater subset.}
\resizebox{\textwidth}{!}{
\begin{tabular}{l l c c c c c c c}
\toprule
Method & Training Set & REL$\downarrow$ & SQ REL$\downarrow$ & RMSE$\downarrow$ & LOG RMSE$\downarrow$ & A1$\uparrow$ & A2$\uparrow$ & A3$\uparrow$ \\
\midrule
\multicolumn{9}{c}{\textbf{Zero-Shot}} \\
\midrule
LEAStereo \cite{cheng2020hierarchical}         &Scene Flow  & 0.1099 & 1.3898 & 4.5610 & 0.2063 & 0.8929 & 0.9512 & 0.9761 \\
PSMNet \cite{chang2018pyramid}                 &Scene Flow  & 0.0884 & 0.8699 & 3.9721 & 0.1804 & 0.9122 & 0.9627 & 0.9804 \\
AANet \cite{xu2020aanet}                       &Scene Flow  & 0.6096 & 8.3687 & 13.0542 & 0.9903 & 0.2598 & 0.3451 & 0.3888 \\
GwcNet \cite{guo2019group}                     &Scene Flow  & 0.1013 & 1.2965 & 4.1829 & 0.1855 & 0.9085 & 0.9612 & 0.9801 \\
ACVNet \cite{xu2022attention}                  &Scene Flow  & 0.0970 & 1.1335 & 3.9985 & 0.1803 & 0.9063 & 0.9612 & 0.9813 \\
RAFT-Stereo \cite{lipson2021raft}              &Scene Flow  & 0.0814 & 0.7342 & 4.0423 & 0.1703 & 0.9030 & 0.9612 & 0.9832 \\
HSMNet \cite{zhao2023high}                      &Scene Flow  & 0.9856 & 12.3768 & 15.2865 & 4.5961 & 0.0000 & 0.0000 & 0.0000 \\
TiO-Depth \cite{zhou2023two}                   &KITTI2012  & 0.7194 & 8.6479 & 13.4635   & 1.6967  & 0.0053 & 0.0096 & 0.0550   \\
FoundationStereo \cite{wen2025foundation}      & FoundationStereo dataset & 0.0542& 0.6701& 2.9644& \textbf{0.1358}& 0.9302& 0.9701& 0.9779\\
Stereo Anywhere \cite{bartolomei2025stereo}    &Scene Flow  & 0.0592 & \textbf{0.5098} & 3.1572 & 0.1544 & 0.9442 & 0.9787 & \textbf{0.9889} \\
CREStereo \cite{li2022practical}            &ETH3D    & 2.5746       &9.8789         &8.4526         &5.1297        &0.4890        &0.5732        &0.7001 \\
\midrule
\textbf{StereoAdapter (Ours)} & UW-StereoDepth-40K & \textbf{0.0527} & 0.5167 & \textbf{2.8947} & 0.1371 & \textbf{0.9467} & \textbf{0.9801} & 0.9873\\
\midrule
\multicolumn{9}{c}{\textbf{Fine-Tuning}} \\
\midrule
IGEV-Stereo \cite{xu2023iterative}             & 5 Datasets$^*$ + TartanAir & 0.1009 & 1.6475 & 4.7107 & 0.1909 & 0.8913 & 0.9515 & 0.9764 \\
Selective IGEV \cite{wang2024selective}        & 5 Datasets$^*$ + TartanAir & 0.1225 & 1.5155 & 4.8742 & 0.2123 & 0.8545 & 0.9375 & 0.9713 \\
GMStereo \cite{xu2023unifying}                 & 5 Datasets$^*$ + TartanAir & 0.1561 & 2.2275 & 5.9224 & 0.2432 & 0.8362 & 0.9252 & 0.9651 \\
\midrule
\textbf{StereoAdapter (Ours)} & TartanAir  &  0.0519 & 0.5041 & 2.8341 & 0.1330 & 0.9489 & 0.9823 & 0.9897\\
\textbf{StereoAdapter (Ours)} & UW-StereoDepth-40K + TartanAir & \textbf{0.0512} & \textbf{0.4987} & \textbf{2.7834} & \textbf{0.1312} & \textbf{0.9512} & \textbf{0.9836} & \textbf{0.9904}\\
\bottomrule
\end{tabular}}
\label{tab:maintable1}
\end{table*}

\subsection{Implementation Details}
\paragraph{Visual augmentation}
We employ the pre-trained MobileIE \cite{yan2025mobileie} as our visual enhancement module to process underwater images and effectively remove water-induced effects. Underwater image degradation is fundamentally a physical problemdue to the selective absorption and scattering of light by water medium, raw images typically exhibit severe color shifts, low contrast, and blurred details. These low-level distortions directly impair feature extraction and recognition performance in subsequent high-level vision tasks. Therefore, applying targeted image restoration at the input stage is essential, as it can recover visual quality without altering semantic content, providing more reliable inputs for downstream tasks. The MobileIE features a lightweight architecture with fast inference speed and delivers stable, consistent enhancement results during both training and testing phases. More importantly, its low computational overhead and memory footprint make it particularly well-suited for deployment on resource-constrained mobile robotic platforms, meeting the real-time visual processing requirements of underwater ROVs.

\paragraph{Model Details}
The encoder of our model is initialized with pre-trained DepthAnything v2-B \cite{yang2024depth} weights. We employ a two-stage training strategy: 20 epochs, and followed by 40 epochs for stereo depth estimation. A constant learning rate of $1 \times 10^{-4}$ is used with the AdamW optimizer and a batch size of 8. Our model is trained on the proposed UW-StereoDepth-40K dataset, with evaluation performed on the TartanAir~\cite{wang2020tartanair} underwater subset and SQUID~\cite{berman2020underwater} datasets. All experiments are conducted on an Intel Xeon Platinum 8469C CPU at 2.60GHz, with a single NVIDIA L40 48GB GPU and 64GB of RAM.

\subsection{Main Results}
The \textit{5 Datasets$^*$} refers to a combination of five commonly used stereo matching datasets for training, which are Scene Flow \cite{mayer2016large}, Sintel \cite{butler2012naturalistic}, ETH3D \cite{schoeps2017cvpr}, InStereo2K \cite{Bao2020InStereo2KAL}, CREStereo \cite{li2022practical}.

Our experiments demonstrate that the proposed StereoAdapter, trained on the UW-StereoDepth-40K dataset, achieves consistent improvements over existing stereo matching methods across both the TartanAir Underwater part and SQUID benchmarks. As summarized in Tables \ref{tab:maintable1} and \ref{tab:maintable2}, our approach delivers state-of-the-art zero-shot performance and achieves further gains when fine-tuned with TartanAir, and UW-StereoDepth-40K.

As shown in Table \ref{tab:maintable1}, StereoAdapter significantly outperforms existing methods on the TartanAir Underwater subset. In the zero-shot setting, our model achieves the lowest REL (0.0527) and RMSE (2.8947), along with the highest accuracy at A1 (94.67\%). When fine-tuned with TartanAir, StereoAdapter further improves across all metrics, reducing RMSE to 2.7834 and achieving 95.12\% for A1, 98.36\% for A2, and 99.04\% for A3. These results demonstrate both the robustness of our adapter design and the value of UW-StereoDepth-40K as a pretraining dataset for underwater applications. 

StereoAdapter attains the best overall performance as shown in Table \ref{tab:maintable2}, achieving an RMSE of 1.8843, which is further reduced to 1.8621 when combined with TartanAir fine-tuning. The model also delivers the highest accuracy across all $\delta$ thresholds, reaching 94.13\% (A1), 97.48\% (A2), and 98.52\% (A3). Compared to other baseline methods, StereoAdapter demonstrates superior generalization capability in challenging underwater conditions. Collectively, these findings highlight the critical role of the UW-StereoDepth-40K dataset in enabling robust zero-shot generalization and fine-tuned performance for underwater stereo depth estimation.

\begin{table*}[t]
\centering

\caption{Zero-shot evaluation on SQUID dataset.}
\resizebox{\textwidth}{!}{
\begin{tabular}{l l c c c c c c c}
\toprule
Method &Training Set & REL$\downarrow$ & SQ REL$\downarrow$ & RMSE$\downarrow$ & LOG RMSE$\downarrow$ & A1$\uparrow$ & A2$\uparrow$ & A3$\uparrow$ \\
\midrule
LEAStereo \cite{cheng2020hierarchical}      &Scene Flow    & 0.5574 & 3.9434  & 5.4659  & 0.4335 & 0.6512 & 0.8042 & 0.8869 \\
PSMNet \cite{chang2018pyramid}              &Scene Flow    & 0.5182 & 7.1404  & 4.9186  & 0.5902 & 0.7139 & 0.7999 & 0.8311 \\
AANet \cite{xu2020aanet}                    &Scene Flow    & 7.4801 & 314.1577& 34.7612 & 1.8994 & 0.0602 & 0.1087 & 0.1570 \\
GwcNet \cite{guo2019group}                  &Scene Flow    & 0.2294 & 1.2275  & 3.0003  & 0.3799 & 0.7423 & 0.8517 & 0.9005 \\
ACVNet \cite{xu2022attention}               &Scene Flow    & 1.6030 & 65.6518 & 10.3828 & 0.7293 & 0.7019 & 0.7925 & 0.8321 \\
RAFT-Stereo \cite{lipson2021raft}           &Scene Flow    & 0.0831 & 0.6946  & 1.9625  & 0.1441 & 0.9235 & 0.9634 & 0.9835 \\
HSMNet \cite{zhao2023high}                  &Scene Flow    & 0.9772 & 7.2766  & 8.2301  & 4.0887 & 0.0000 & 0.0000 & 0.0000 \\
CREStereo \cite{li2022practical}            &ETH3D    & 2.5746       &9.8789         &8.4526         &5.1297        &0.4890        &0.5732        &0.7001        \\
IGEV-Stereo \cite{xu2023iterative}          & 5 Datasets$^*$ + TartanAir & 0.0932 & 1.4685 & 2.4741 & 0.1523 & 0.9346 & 0.9712 & 0.9820 \\
Selective IGEV \cite{wang2024selective}     & 5 Datasets$^*$ + TartanAir & 0.0960 & 0.9617  & 1.9268  & 0.1665 & 0.9171 & 0.9555 & 0.9720 \\
GMStereo \cite{xu2023unifying}              & 5 Datasets$^*$ + TartanAir & 3.3442 & 140.3211 & 18.7829 & 1.0219 & 0.5300 & 0.6076 & 0.6578 \\
TiO-Depth \cite{zhou2023two} &KITTI2012  & 1.3154 & 11.6828 & 7.0930   & 0.8121  & 0.1753 & 0.3346 & 0.5133   \\
FoundationStereo \cite{wen2025foundation}   &FoundationStereo dataset    & 0.1095 & 0.7012  & 2.2510  & 0.1584 & 0.8995 & 0.9433 & 0.9501 \\
Stereo Anywhere \cite{bartolomei2025stereo} &Scene Flow    & 0.0952 & 1.1017  & 2.4317  & 0.1586 & 0.9179 & 0.9605 & 0.9763 \\
\midrule
\textbf{StereoAdapter (Ours)}& UW-StereoDepth-40K & 0.0806 & 0.7082 & 1.8843 & 0.1469 & 0.9413 & 0.9748 & 0.9852\\
\textbf{StereoAdapter (Ours)} & UW-StereoDepth-40K + TartanAir & \textbf{0.0795} & \textbf{0.6823} & \textbf{1.8621} & \textbf{0.1398} & \textbf{0.9428} & \textbf{0.9761} & \textbf{0.9867}\\
\bottomrule
\end{tabular}}
\label{tab:maintable2}
\end{table*}

\begin{figure*}[t]
    \centering
    \includegraphics[width=\textwidth]{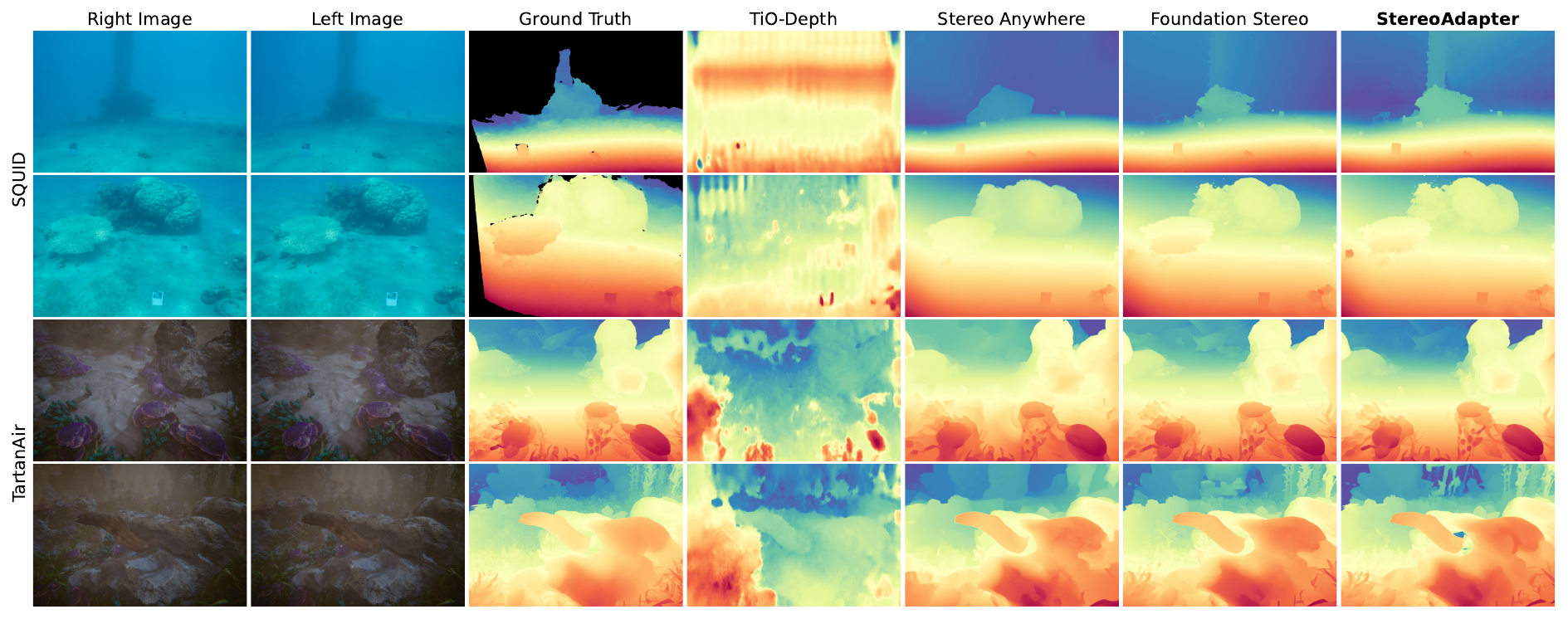}
    \caption{{\textbf{Visualization results} of Stereo Depth Estimation Methods on SQUID and TartanAir.} }
    \label{fig:result_sample}
\end{figure*}

As shown in Figure \ref{fig:result_sample}, StereoAdapter generates substantially more accurate and visually coherent depth maps than baseline methods (for example, better scale estimation for far range areas).

\subsection{Real World Evaluation}
\begin{table}[t]
\centering
\caption{\textbf{Real World Evaluation} on BlueROV2.}
\label{tab:realworld_result}
\resizebox{\columnwidth}{!}{
\begin{tabular}{l c c c c c}
\toprule
Method & REL$\downarrow$ & SQ REL$\downarrow$ & RMSE$\downarrow$ & LOG RMSE$\downarrow$ & A1$\uparrow$ \\
\midrule
Stereo Anywhere \cite{bartolomei2025stereo} & 0.0905 & 1.0467 & 2.5101 & 0.1507 & 0.9120\\
FoundationStereo \cite{wen2025foundation}   & 0.1040 & 0.7363 & 2.1385 & 0.1505 & 0.8961\\
TiO-Depth \cite{zhou2023two}                & 1.5697 & 13.1487 & 6.7584 & 0.8715 & 0.1525\\
\textbf{StereoAdapter (Ours)}               & \textbf{0.0856} & \textbf{0.6482} & \textbf{1.9690} & \textbf{0.1428} & \textbf{0.9478} \\
\bottomrule
\end{tabular}}
\end{table}

\begin{figure}[t]
    \centering
    \includegraphics[width=\columnwidth]{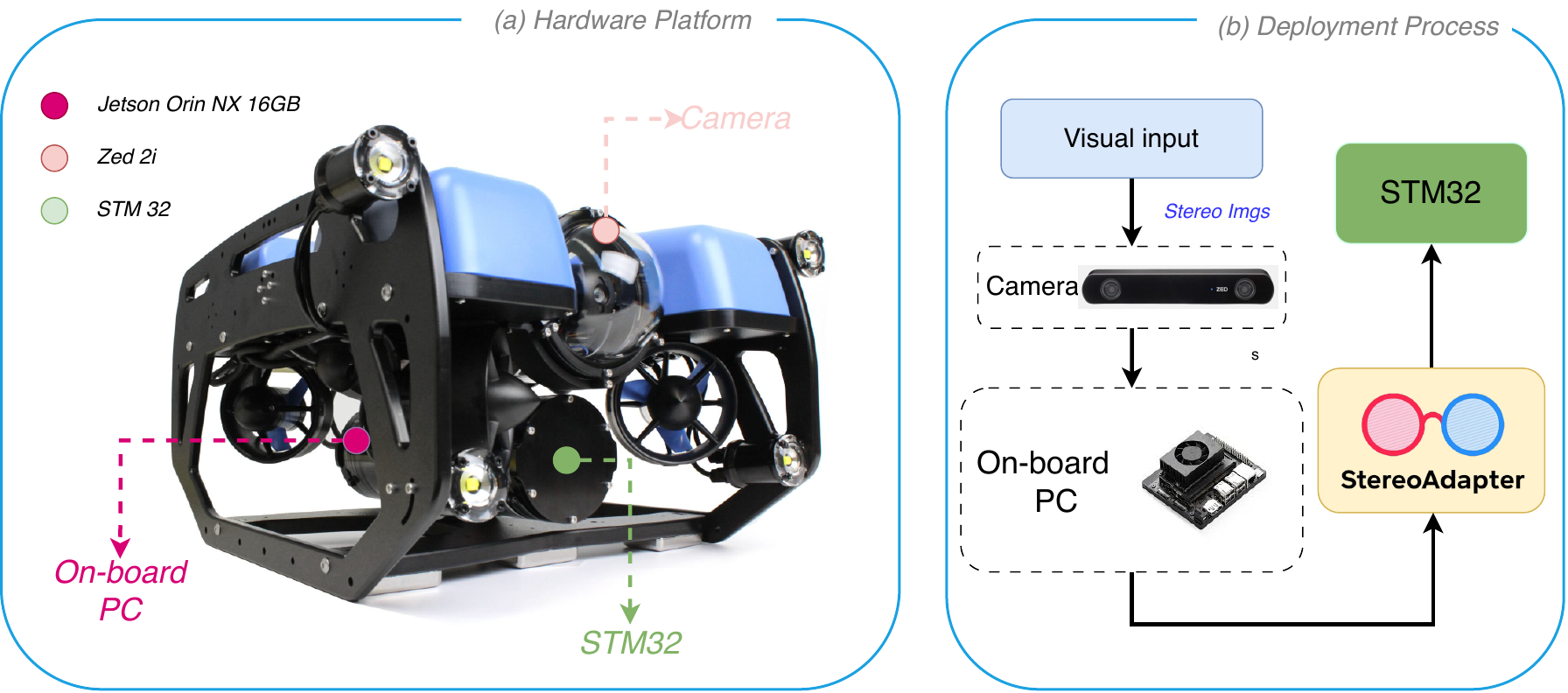}
    \caption{\textbf{Hardware and Evaluation Pipeline} for Real-World Experiments.}
    \label{fig:rovsetup}
\end{figure}

\paragraph{Platform} As shown in Figure \ref{fig:rovsetup}, we employ a BlueROV2 as the experimental vehicle. Low-level actuation is handled by an STM32 controller, while onboard perception and inference run on a Jetson Orin NX (16GB). The robot is equipped with a stereo camera Zed 2i; the left/right sensors are triggered by the camera’s built-in sync circuitry. All sensing and logging are performed onboard the Orin NX.

\paragraph{Environment and Obstacles}
Experiments are performed in the same indoor rectangular water tank. To emulate underwater obstacle avoidance, we arrange glass cups and rocks of varying shapes to form three distinct obstacle environments within this single scene: dispersed, side-by-side, and clustered. For each environment, the robot is teleoperated along three different pre-defined motion trajectories, and we record one synchronized stereo sequence per trajectory. All data are logged onboard the Orin NX.

\paragraph{AprilTag Layout and Reference Geometry}
Before the experiments, we pre-constructed a metrically scaled 3D mesh of the scene. During the underwater runs, camera poses were estimated from AprilTag (family 16h5) detections via PnP and aligned to the mesh. For each frame, we then rendered a reference depth map in the left-camera coordinate frame with standard visibility handling, masking pixels without valid mesh hits. These maps serve as the ground truth for evaluating our predicted depth across all three obstacle layouts. We evaluate only on sequences with dense mesh coverage, ensuring effectively dense and reliable ground-truth depth in our demos.

\paragraph{Protocol}
We evaluate all methods on the same indoor tank described above. Within this single scene we realize three obstacle environments (dispersed, side-by-side, clustered) using glass cups and rocks, and for each environment we teleoperate the robot along three distinct motion trajectories, yielding nine synchronized stereo sequences in total. All approaches consume the identical rectified stereo pairs at the same input resolution with the same pre-processing. Where a method outputs disparity, we convert to metric depth by using the calibrated focal length 
$f$ and baseline $b$. Unless noted otherwise, evaluation is performed in the left-camera coordinate frame.

\paragraph{Masking and evaluation scope}
We evaluate against the per-frame reference depth rendered from the metrically scaled scene mesh described above. During evaluation, depth is compared in the left-camera coordinate frame; pixels without a valid z-buffer hit in the rendered reference are masked and excluded. Metrics are reported over valid pixels only, and we restrict reporting to sequences with dense reference coverage.

\paragraph{Metrics}
We evaluate using the standard depth metrics shown in Table~\ref{tab:realworld_result}: Absolute Relative Error (REL), Squared Relative Error (SQ\,REL), RMSE, LOG RMSE, and accuracy $\mathbf{A1}$ with $\delta<1.25$. Metrics are computed per frame over \emph{valid} pixels only (as defined by the reference-depth mask) and averaged across the nine sequences (three environments $\times$ three trajectories). Lower is better for REL/SQ\,REL/RMSE/LOG RMSE; higher is better for A1.

\paragraph{Baselines}
We compare against representative stereo approaches: {Stereo Anywhere}~\cite{bartolomei2025stereo}, {FoundationStereo}~\cite{wen2025foundation}, and the stereo branch of {TiO-Depth}~\cite{zhou2023two}. Our method StereoAdapter is evaluated under identical input settings and without augmentation of the test time.

\subsection{Ablation Study}

\begin{table}[t]
  \centering
  \caption{Ablation on \textbf{recurrent refinement module}.}
  \label{tab:ablate_gru}
  \resizebox{\columnwidth}{!}{
  \begin{tabular}{l c c c c }
    \toprule
    GRU layers & Hidden Dimension & Number of Iteration  & REL$\downarrow$ & RMSE$\downarrow$ \\
    \midrule
    4  & 128 & 32 & 0.049 & 2.614  \\
    3 & 256 & 32 & 0.048 & 2.625 \\
     3 & 128 & 64 & 0.533 & 2.8654 \\
     \rowcolor{yellow!25} 3 & 128  & 32 & 0.051&  2.783  \\
     2 & 128 & 32 & 0.595 & 3.024 \\
    \bottomrule
  \end{tabular}}
\end{table}

\begin{table}[t]
  \centering
  \caption{Ablation on \textbf{Dynamic LoRA} setups.}
  \label{tab:ablate_lora}
  \resizebox{\columnwidth}{!}{
  \begin{tabular}{l c c c c}
    \toprule
    Rank  & $\kappa$ Threshold& Dense Epoch Ratio& REL$\downarrow$ & RMSE$\downarrow$  \\
    \midrule
    16 & 0.005 & 0.5 & 0.077 & 3.214  \\
    16 & 0.005 & 0.45 & 0.074 & 3.105 \\
    \rowcolor{yellow!25} 16 & 0.01 & 0.45 & 0.049 & 2.783 \\
    32 & 0.005 & 0.5 & 0.049 & 2.814 \\
    32 & 0.01 & 0.5 & 0.054 & 2.744 \\     
    \bottomrule
  \end{tabular}}
\end{table}

\begin{table}[t]
  \centering
  \caption{Ablation on Training strategy}
  \label{tab:ablate_optim}
  \resizebox{\columnwidth}{!}{
  \begin{tabular}{l c c c c c c c c c}
    \toprule
    Batch Size & Learning Rate & Stage One Epochs & Stage Two Epochs & REL$\downarrow$/RMSE$\downarrow$  \\
    \midrule
     4   & $1 \times 10^{-4}$ & 20 & 20 & 0.0711/3.183 \\
    4   & $2 \times 10^{-4}$ & 30 & 60 & 0.0667/2.951 \\
    \rowcolor{yellow!25} 8 & $1 \times 10^{-4}$ & 20 & 40 & 0.051/2.783 \\
     8   & $2 \times 10^{-4}$ & 30 & 30 & 0.054/2.842 \\
     16 & $1 \times 10^{-4}$ & 20 & 20 & 0.052/2.805 \\
    \bottomrule
  \end{tabular}}
\end{table}

We conduct comprehensive ablation experiments to evaluate the impact of key design choices in our framework. Specifically, we investigate: (i) the recurrent refinement module configuration, (ii) the Dynamic LoRA adaptation strategy, and (iii) critical training hyperparameters.

Table~\ref{tab:ablate_gru} shows the ablation results for the recurrent refinement module. We vary the number of GRU layers, hidden dimensions, and refinement iterations. The results demonstrate that deeper networks significantly improve performance, with the 4-layer architecture achieving the best results. Our final configuration with \textbf{3} GRU layers, \textbf{128} hidden dimensions, and \textbf{32} iterations balances performance with computational efficiency.

Table~\ref{tab:ablate_lora} reports the results for different Dynamic LoRA configurations. We examine the adapter rank, $\kappa$ threshold for rank scheduling, and the dense epoch ratio. Our configuration employs a rank of \textbf{16} with \textbf{$\kappa$ = 0.01} and \textbf{45\%} dense epochs; it has the best performance on RMSE and REL. This LoRA configuration provides both high performance and training efficiency.

\begin{figure*}[t]
    \centering
    \includegraphics[width=\textwidth]{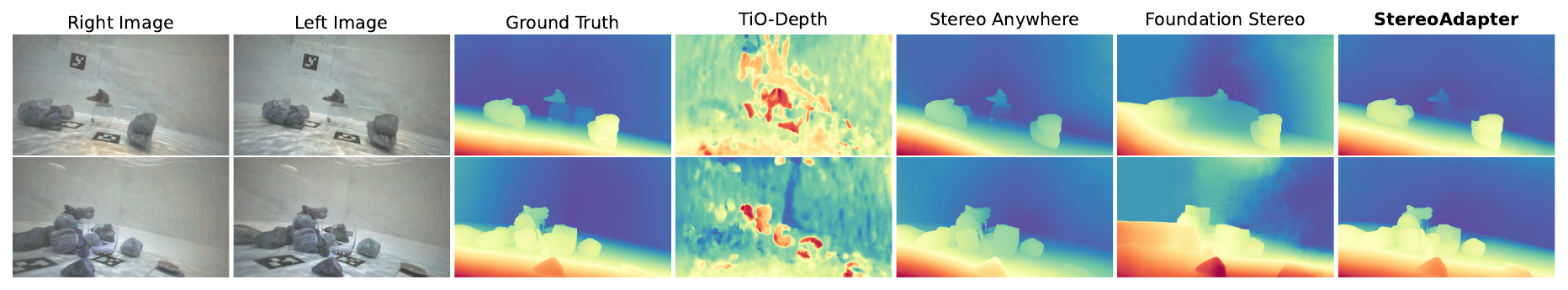}
    \caption{{Generalization of Stereo Methods to Real-World Experiments} }
    \label{fig:real_world_sample}
\end{figure*}

Table~\ref{tab:ablate_optim} examines the influence of training hyperparameters including batch size, learning rate, and training epochs. The experiments reveal that moderate batch sizes (8-16) consistently outperform smaller batches, while a learning rate of $1\times10^{-4}$ ensures more stable convergence compared to $2\times10^{-4}$. Notably, Multistage training epoch allocation (20 epochs for stage one, 40 for stage two) allows the model to first learn monocular features before learning on stereo domain, yielding optimal results with batch size 8 (REL=0.051, RMSE=2.783). Our final configuration employs a batch size of \textbf{8}, learning rate of $\boldsymbol{1\times10^{-4}}$, \textbf{20} epochs for first stage, and \textbf{40} epochs for second stage. Overall, our quantitative evaluation confirms that StereoAdapter outperforms existing state-of-the-art methods across multiple metrics while maintaining good efficiency.

\section{Test-Time Efficiency}
\label{sec:test_time}

We evaluate on an on-board {Jetson Orin NX 16GB} in {MaxN} mode with TensorRT, batch size~1, and input resolution {$640{\times}320$}. All methods use authors’ official implementations with identical pre/post-processing. We report per-frame end-to-end latency in milliseconds (ms).

\begin{table}[t]
\centering
\caption{Average per-frame \textbf{inference latency (ms)} on Jetson Orin NX @ $640{\times}360$, BS=1.}
\label{tab:orin_nx_latency}
\vspace{2pt}
\begin{tabular}{l c}
\toprule
\textbf{Method} & \textbf{On-board (ms)} \\
\midrule
FoundationStereo\cite{wen2025foundation} & {1815} \\
Stereo Anywhere \cite{bartolomei2025stereo} & {1440} \\
StereoAdapter (Ours) & \textbf{1113} \\
\bottomrule
\end{tabular}
\vspace{-0.5cm}
\end{table}

FoundationStereo is the slowest on Orin NX , consistent with its heavy transformer backbone and comprehensive cost aggregation; Stereo Anywhere improves to 1440\,ms by using a RAFT-style recurrent module; its budget is dominated by running DepthAnything-L for twice and the {two} monocular passes, and a 3D convolution module for feature fusion. In contrast, {StereoAdapter} is the fastest at 1113\,ms: it leverages a LoRA-adapted DepthAnything-B encoder only for feature extraction, yielding a 327\,ms speedup over Stereo Anywhere and a 702\,ms speedup over FoundationStereo on the same board.

\section{Limitation and Future Work}

Our approach focuses on a lightweight decoder with LoRA-aided specialization and synthetic supervision. This yields strong efficiency and robustness in our evaluated settings, but several limitations remain. \textit{(i) Decoder capacity vs.\ context:} the current RAFT-style recurrent head emphasizes local matching with short-range memory; under severe turbidity, non-Lambertian highlights, and large textureless spans, global reasoning is still limited. \textit{(ii) Synthetic supervision (UE):} while Unreal Engine simulation substantially reduces labeling cost, its data distribution remains narrower and more factorized than real-world underwater imagery. In particular, the pipeline provides limited domain breadth and scene co-occurrence statistics, and simplifies correlations among appearance, geometry, and control; important phenomena (participating-media multiple scattering, polarization effects, dynamic turbidity/particles, sensor/ISP idiosyncrasies, rolling-shutter and lens aberrations) are only partially captured, leaving a measurable sim-to-real gap.

Future work will broaden both data and model. On the data side, we plan a unified training recipe that couples UE procedural generation with coverage-driven domain randomization and physics-based participating-media rendering, calibrated to measured sensors and scene spectra; we will also explore self-training on unlabeled real sequences with confidence filtering to close residual gaps. On the model side, we will investigate unified multi-task objectives (e.g., stereo, depth, normals) to absorb broader cross-task statistics, alongside longer-context, linear-time decoders (such as Mamba/RWKV) and temporal/multi-view extensions with explicit uncertainty estimation and quantization-aware deployment on embedded platforms.

\section{Conclusion}
In this work, we present \textbf{StereoAdapter}, a parameter-efficient self-supervised framework that integrates a LoRA-adapted monocular foundation encoder with a recurrent stereo refinement module for underwater depth estimation. Our method addresses the challenge of adapting large vision models to severe underwater domain shifts by leveraging dynamic LoRA adaptation and pre-training on the synthetic \textbf{UW-StereoDepth-40K} dataset. Comprehensive evaluations demonstrate StereoAdapter achieves 6.11\% improvement on TartanAir and 5.12\% on SQUID compared to state-of-the-art methods, while real-world deployment on BlueROV2 confirms robust performance with favorable latency. These advancements provide practical solutions for AUV navigation, infrastructure inspection, and marine ecology monitoring, contributing to safer autonomous underwater operations.

\section{Appendix: Visualization}

\begin{figure*}[t]
    \centering
    \includegraphics[width=0.90\textwidth]{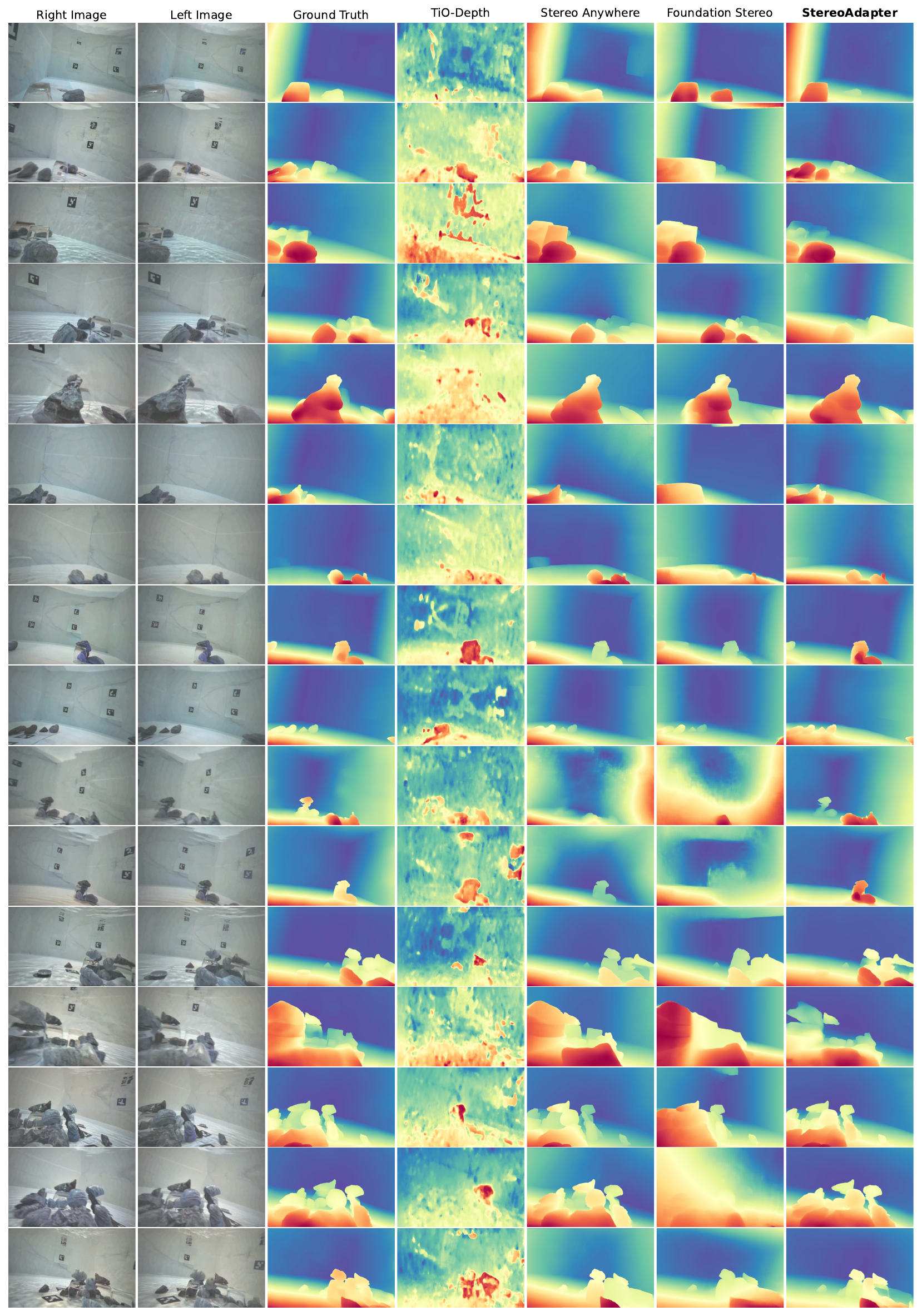}
    \caption{Comparison results from real-world experiments}
    \label{fig:extra_realworld}
\end{figure*}

\paragraph{Real world evaluation} To assess real-world performance, we present qualitative results on BlueROV2 across multiple key frames and compare StereoAdapter with captured ground truth and competing baselines (Fig.~\ref{fig:extra_realworld}). In particular, StereoAdapter consistently produces sharper and more geometrically faithful depth maps than competing baselines, especially in regions with textureless water and reflective surfaces where traditional methods often fail.

\begin{figure*}[t]
    \centering
    \includegraphics[width=0.90\textwidth]{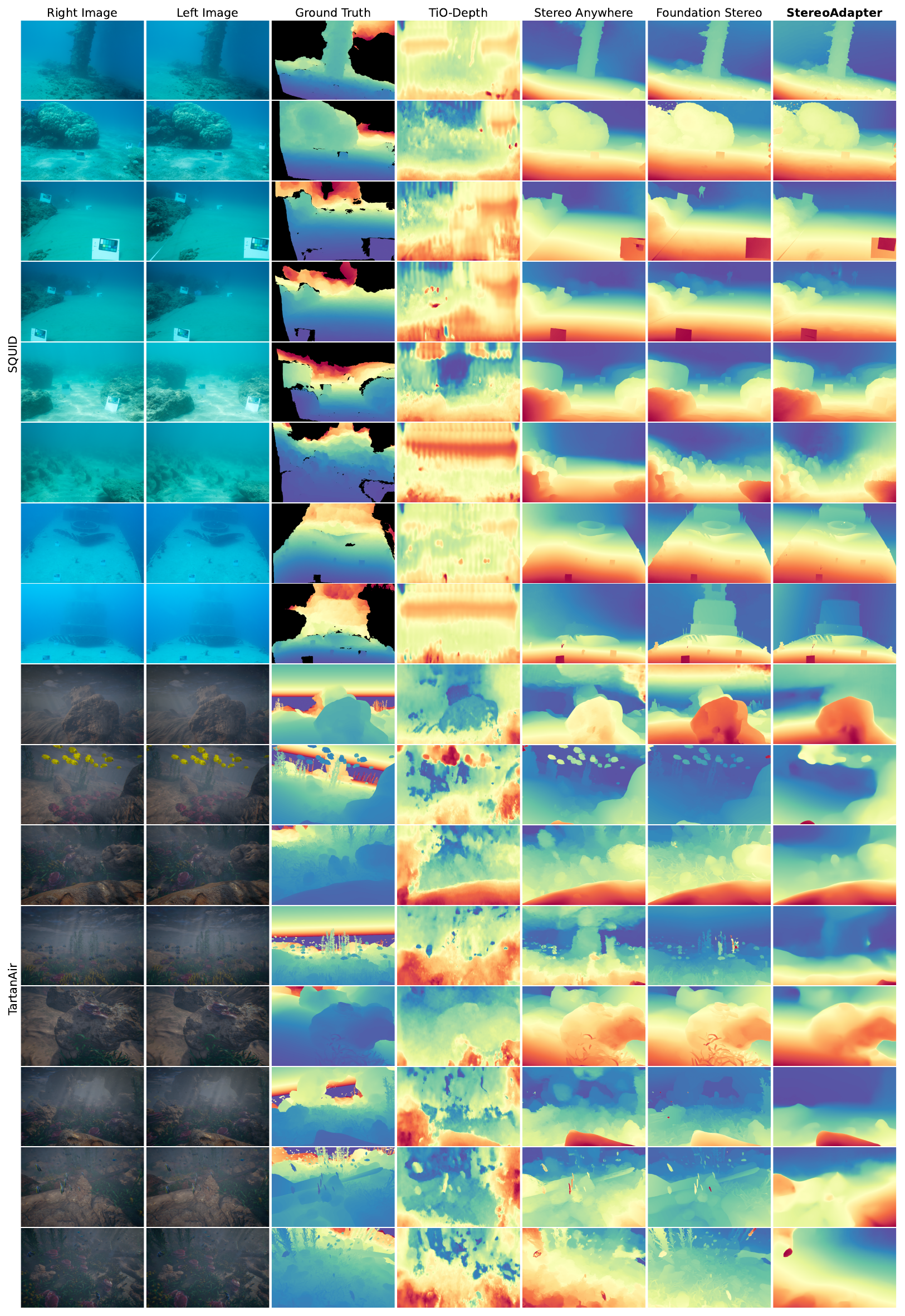}
    \caption{Comparison results from TartanAir and SQUID}
    \label{fig:datasets}
\end{figure*}

\paragraph{Benchmark evaluation} We further report additional qualitative visualizations on TartanAir and SQUID (Fig.~\ref{fig:datasets}). Results on TartanAir and SQUID further highlight its ability to generalize across diverse synthetic-to-real domain shifts, capturing fine-grained scene structure with fewer artifacts. 

\begin{figure*}[t]
    \centering
    \includegraphics[width=0.90\textwidth]{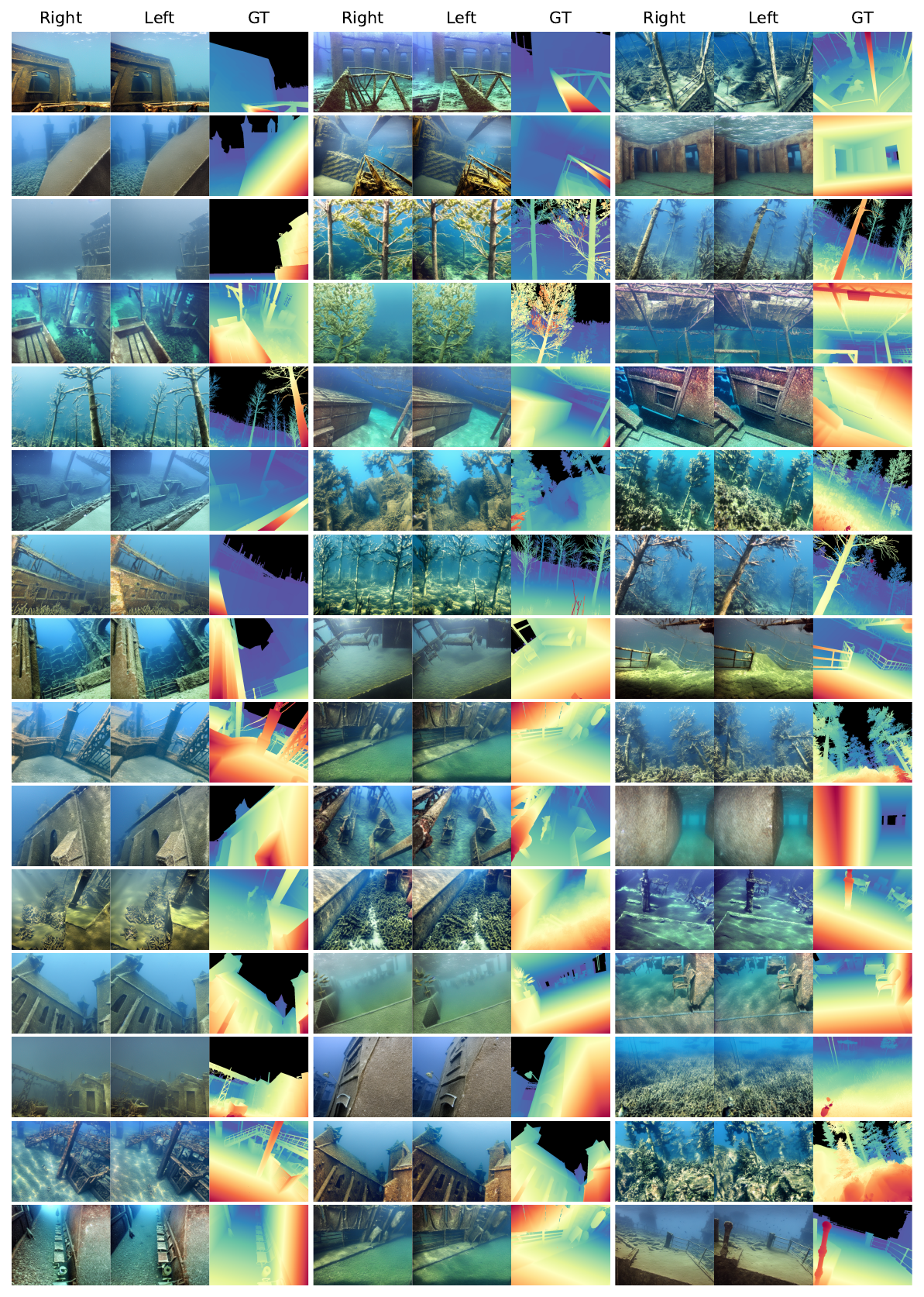}
    \caption{Synthetic underwater images from UW-StereoDepth-40K}
    \label{fig:synthetic_images}
\end{figure*}

\paragraph{UW-StereoDepth-40K visualization} Beyond these qualitative comparisons, we showcase representative image-depth pairs from our proposed UW-StereoDepth-40K dataset (Fig.~\ref{fig:synthetic_images}). The examples from UW-StereoDepth-40K demonstrate the dataset’s variety in viewpoint, illumination, and scene composition, underscoring its value as a large-scale resource for training underwater depth estimation models.


\clearpage
\clearpage
% \bibliographystyle{IEEEtran}
% \bibliography{IEEEabrv}

\end{document}